\documentclass{article}
\usepackage{spconf,amsmath,epsfig}
\usepackage{multirow}
\let\OLDthebibliography\thebibliography
\renewcommand\thebibliography[1]{
  \OLDthebibliography{#1}
  \setlength{\parskip}{0pt}
  \setlength{\itemsep}{0pt plus 0.3ex}
}
\usepackage{amsthm,amsmath,amssymb}
\usepackage{mathrsfs}
\usepackage[hidelinks]{hyperref}
\begin{document}\sloppy

\def\x{{\mathbf x}}
\def\L{{\cal L}}

\title{Support-set based Multi-modal Representation Enhancement \\for Video Captioning}
%
%
\name{Xiaoya Chen$^{1}$, Jingkuan Song$^{1,*}{\thanks{* Corresponding author.}}$, Pengpeng Zeng$^{1}$, Lianli Gao$^{1}$, Heng Tao Shen$^{1}$
\thanks{This work is supported by National Key Research and Development Program of China (No. 2018AAA0102200), the National Natural Science Foundation of China (Grant No. 62122018, No. 61772116, No. 61872064), Sichuan Science and Technology Program (Grant No.2019JDTD0005).}}
\address{$^{1}$University of Electronic Science and Technology of China, China\\ 
\{ccchenxiaoya,jingkuan.song,is.pengpengzeng\}@gmail.com\\lianli.gao@uestc.edu.cn; shenhengtao@hotmail.com}

\maketitle

\begin{abstract}
Video captioning is a challenging task that necessitates a thorough comprehension of visual scenes. Existing methods follow a typical one-to-one mapping, which concentrates on a limited sample space while ignoring the intrinsic semantic associations between samples, resulting in rigid and uninformative expressions. To address this issue, we propose a novel and flexible framework, namely Support-set based Multi-modal Representation Enhancement (\textbf{SMRE}) model, to mine rich information in a semantic subspace shared between samples. Specifically, we propose a \textbf{Support-set Construction (SC)} module to construct a support-set to learn underlying connections between samples and obtain semantic-related visual elements. During this process, we design a \textbf{Semantic Space Transformation (SST)} module to constrain relative distance and administrate multi-modal interactions in a self-supervised way. Extensive experiments on MSVD and MSR-VTT datasets demonstrate that our SMRE achieves state-of-the-art performance. Our code is released at \href{https://github.com/SMRE-CV/SMRE}{https://github.com/SMRE-CV/SMRE}.
\end{abstract}
\section{Introduction}
\label{sec:intro}

Video captioning aims to generate accurate and diverse language descriptions for videos automatically. Compared to the image captioning task, videos contain more spatio-temporal information, more details, and richer semantics. Therefore, handling such rich potential semantics rationally and improving semantic representation is a significant challenge for this topic.

\begin{figure}[t]
\centering
\includegraphics[width=1.0\linewidth]{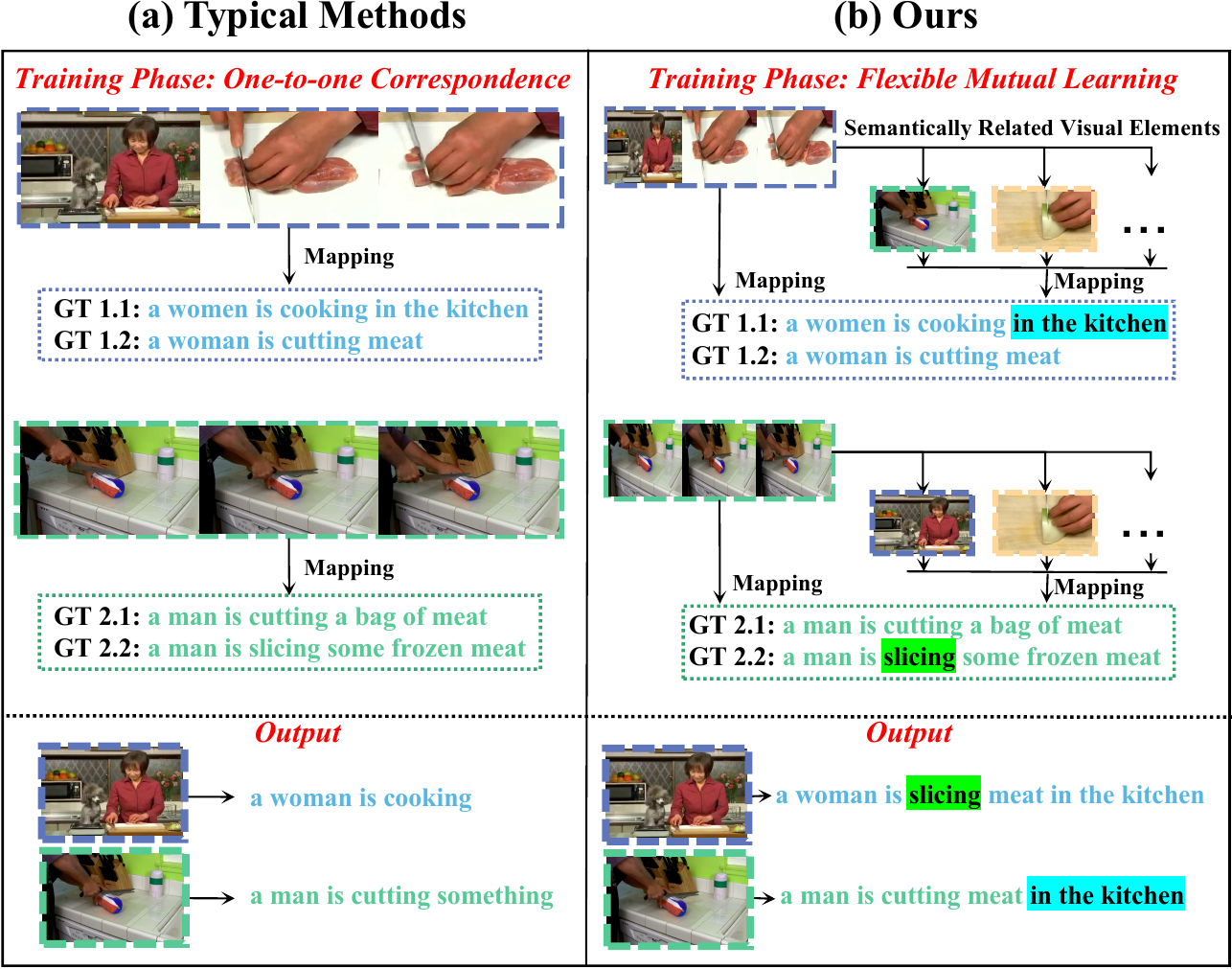}
\caption{An illustration of the differences between traditional methods (a) and ours (b) in training rules and outcomes. GT denotes the ground truth annotations. Typical methods follow a one-to-one mapping to obtain plain statements, while our method establishes a more flexible mapping approach, yielding richer semantic expressions.}
\label{fig:res}
\end{figure}

For better semantic representations, existing works adopt grid features to capture more spatial information and build message interactions within frames \cite{chen2019motion, chen2021motion}. Another perspective is to introduce a pretrained object detector to acquire object guidances for features fusion \cite{pan2020spatio, zheng2020syntax}. However, these typical methods, as shown in Figure 1(a), follow a strict one-to-one mapping from a video to a caption in the training phase. This rigid mapping process concentrates on its own sample space and ignores inherent associations between samples (e.g., videos may contain analogous materials), resulting in restricted and uninformative expressions. As demonstrated by ``output", even if the videos are highly semantically related, the texts generated are limited to their annotations and lack vividness.

To tackle this problem, we introduce a support-set concept, which constructs a set of auxiliary information and captures connections of inner details, inspired by multi-modal representation learning tasks like cross-modal retrieval \cite{patrick2020support}. We adopt this concept for video captioning to capture semantic-related visual elements in a shared semantic subspace and create flexible mappings between modalities. As depicted in Figure 1 (b), our method develops more reasonable mapping relationships, enabling the generated sentences to gain richer expressions from other semantically related videos. It is shown that the ``cutting'' operation in the picture gives the new expression ``slicing'' and a higher level semantics of ``in the kitchen'' is derived from the element of ``cutting meat.''

In this paper, we propose a Support-set based Multi-modal Representation Enhancement (\textbf{SMRE}) model to exploit information in the semantic subspace shared between samples and improve multi-modal representations. Essentially, our model is based on a typical encoder-decoder framework. On top of that, we obtain a support-set via a \textbf{Support-set Construction (SC)} module and feed it into a new branch that shares the encoder-decoder with the original path to establish flexible mappings. Note that the support-set is not applied in the inference phase to avoid cheating information. Moreover, to model complex semantic relationships in the shared subspace, we introduce a \textbf{Semantic Space Transformation (SST)} module with two extra losses to constrain relative distances in multi-modal semantic space and conduct a self-supervised process from two perspectives: inter-modality and intra-modality, respectively. With these losses, the encoder-decoder structure can learn better semantic representations and generate semantic-enriched captions. To summarize, our contributions lie in three folds: 

1) Compared with conventional one-to-one mappings, we build a novel and flexible mapping framework (\textbf{SMRE}) by developing a support-set concept that captures connections of inner details between samples and obtains semantic-related visual elements.

2) To further constrain semantic relationships, we propose a semantic space transformation (\textbf{SST}) in a self-supervised process from two perspectives: inter-modality and intra-modality.



3) Our SMRE achieves state-of-the-art performance on MSVD and MSR-VTT benchmark datasets. Notably, our model outperforms SOTA by 5.3\% in BLEU-4 and 1.1\% in CIDEr on MSVD, and 1.5\% in BLEU-4 and 0.4\% in CIDEr on MSR-VTT, respectively.

\section{Related Work}
\begin{figure*}[ht]
\label{pipeline}
\centering
\includegraphics[width=0.9\textwidth]{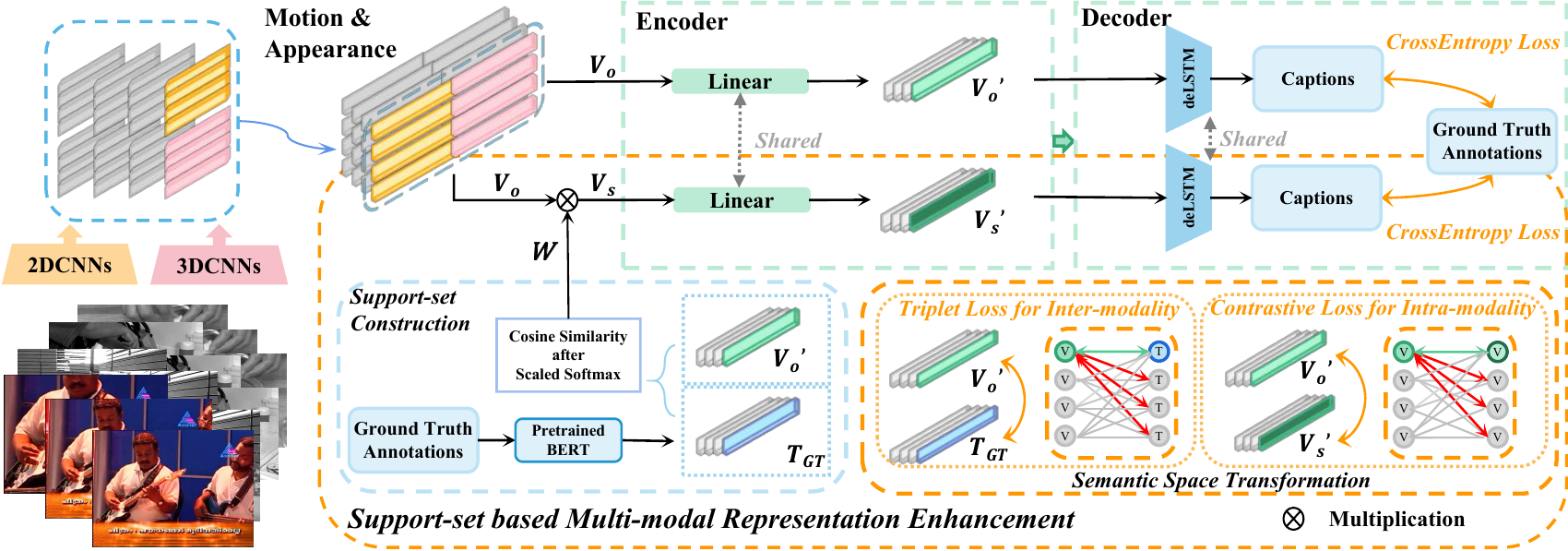}
\caption{Illustration of our proposed SMRE model. Based on the typical framework, we add a new branch beside the original encoder-decoder path. We abstract a support video set via a Support-set Construction (SC) module and map it to the ground truth annotations. Moreover, we design a Semantic Space Transformation (SST) module that incorporates triplet loss and contrastive loss to constrain relative distance and administrate multi-modal interactions in a shared semantic subspace.}
\label{fig:res}
\end{figure*}
\textbf{Video Captioning.} Video captioning is a forerunner in multi-modal tasks, and the concept has undergone extensive development. The majority of early video captioning algorithms relied on specific textual templates and necessitated a huge proportion of hand-crafted linguistic rules \cite{das2013thousand}. With the emergence of deep neural networks, these limitations are addressed by a typical encoder-decoder framework. For instance, Xu \emph{et al.} \cite{xu2015show} proposed a widely-used soft-attention encoder-decoder model, becoming a paradigm for follow-up works. To further improve the performance, various changes have been made in recent years. We divide recent works into three categories: 1) those that continued the original pipeline and focused more on the fusion of modalities or embedding methods\cite{wang2018m3, pei2019memory, aafaq2019spatio, wang2019controllable, zhao2021multi, ryu2021semantic, Gao, openbook}. For example, M$^3$ \cite{wang2018m3} created a shared visual and textual memory to simulate long-term visual-textual dependency and to guide global attention on described targets. 2) those that introduced grid features to capture more grid-level spatial information \cite{chen2019motion, chen2021motion}. MGRMP \cite{chen2021motion} designed a spatial information extraction and aggregation method to utilize fine-grained features. Furthermore, 3) those that introduced the object detector to increase the spatio-temporal interaction between objects \cite{pan2020spatio, zheng2020syntax, tan2020learning}. For example, SAAT \cite{zheng2020syntax} learned actions by simultaneously referring to dynamic and static representations of objects.

In contrast to these approaches, our approach does not introduce external features or human-made language rules but instead builds more flexible mappings by better exploring existing relationships in a shared semantic subspace, allowing the model to learn richer semantic representations. 

\noindent\textbf{Semantic Spatial Distance Transformation.} Semantic gaps have always hindered the process of multi-modal tasks. However, learning the sophisticated multi-modal relations by designing specific architecture is difficult. To tackle this problem, contrastive learning, which is widely used in the field of pre-training, is gradually transferred to particular tasks like cross-modal retrieval \cite{Zeng} and video grounding with a support-set concept \cite{patrick2020support,ding2021support}. Inspired by these works, we introduce contrastive learning methods to the video captioning domain with a support-set architecture to model complex relationships in multi-modal semantic space.

\section{support-set based multi-modal representation enhancement}

In this section, we introduce a Support-set based Multi-modal Representation Enhancement model to mine information in a shared semantic subspace and learn better semantic representations. Figure 2 depicts the pipeline of our SMRE, which extends the encoder-decoder structure with a new input branch. These two paths share the same encoder-decoder. For the encoder (Section 3.1), we add a Support-set Construction (SC) module (Section 3.1.1) and a Semantic Space Transformation (SST) module (Section 3.1.2) with two contrastive learning methods: (1) Triplet Loss for inter-modality interaction; (2) Contrastive Loss for intra-modality interaction. The typical LSTM structure is retained for the decoder (Section 3.2), and the Cross-Entropy loss is used to constrain the overall generating process.

\subsection{Encoder}
In the encoder module, a support-set is first constructed by the SC module and mapped to the ground truth annotations with the new branch. During this process, two contrastive learning methods in the SST module are applied to conduct inter-modality and intra-modality interactions and capture subtle semantic relationships in the shared multi-modal semantic subspace.

\subsubsection{Support-set Construction}
In this section, our support-set is built from raw video features by following two rules: 1) the support-set should own semantic-enriched visual information; 2) the support-set should be acquired from the existing resource without additional data.

To capture precise relationships, we apply corresponding ground truth annotations as guidance. Let $B$ denote the batch size. The Support-set Construction Module computes the cosine similarity $\textbf{S} \in \mathbb{R}^{B\times B}$ between the output of the encoder, i.e., ${\textbf{V}_O}^\prime$, and text embedding $\textbf{T}_{GT}$ by a pretrained language BERT, further processes it with a softmax function and views it as weight parameters $\textbf{W} \in \mathbb{R}^{B\times B}$. Then, the module multiplies $\textbf{W}$ times origin video features $\textbf{V}_{O}$ to get support video feature $\textbf{V}_{S}$, which can be considered as a weighted sum according to the semantic similarity between the input visual-contextual pairs. The operational procedure can be represented as follows:
\begin{equation}
\begin{aligned}
    &\textbf{S}=cosine\_similarity(\textbf{T}_{GT},{\textbf{V}_O}'), \\
    &\textbf{W}=softmax(\theta_{scale}\textbf{S}),\\
    &\textbf{V}_{S}=\textbf{W}\otimes \textbf{V}_{O},
\end{aligned}
\end{equation}
where $\theta_{scale}$ is a constant to make the data more dispersive. The support-set $\textbf{V}_{S}$ is then fed into the same encoder that contains a linear layer and gets the middle state ${\textbf{V}_{S}}'$.
\subsubsection{Semantic Space Transformation}
\textbf{Inter-Modality Interaction.} For inter-modality interaction, we import triplet loss with the use of hard negatives. To provide positive feedback on the learning course of the encoder, we apply this loss between video embedding ${\textbf{V}_O}'$ after the encoder and text embedding $\textbf{T}_{GT}$ after the pretrained BERT to bring the semantic distance between these two modalities closer. 

This triplet loss called Max of Hinges \cite{faghri2017vse++} focuses on hard negatives for training the negatives closest to each training query. Let $P={(\textbf{v}_i,\textbf{t}_i)}_{i=1}^B$ be a set of Video-Text pairs, $s(\textbf{a},\textbf{b})=\frac{\textbf{a}^T\textbf{b}}{\parallel \textbf{a}\parallel\parallel \textbf{b}\parallel}$ be the cosine similarity of vector \textbf{a} and \textbf{b}. If we have a positive pair $\left(\textbf{v},\textbf{t}\right)$, the hardest negatives are given by $\textbf{v}'=argmax_{m\neq v} s\left(\textbf{m},\textbf{t}\right)$ and  $\textbf{t}'=argmax_{n\neq t} s\left(\textbf{v},\textbf{n}\right)$. Following the rules of triplet loss, our loss can be defined as:
\begin{eqnarray}
l_{inter} =\max\limits_{\textbf{t}'}\left[\alpha+s\left(\textbf{v},\textbf{t}'\right)-s\left(\textbf{v},\textbf{t}\right)\right]_+ \nonumber\\
+\max\limits_{\textbf{v}'}\left[\alpha+s\left(\textbf{v}',\textbf{t}\right)-s\left(\textbf{v},\textbf{t}\right)\right]_+,
\end{eqnarray}
where $[\cdot]_+=max(0,\cdot)$. The margin $\alpha$ is set to be 0.2. In this way, the module can reduce the semantic distance between positive V-T pairs while increasing the semantic distance between negative V-T pairs until they reach the margin. Furthermore, the inter-modality gap can be narrowed.

\noindent\textbf{Intra-Modality Interaction.} For intra-modality interaction, we apply the contrastive loss to capture relationships between ${\textbf{V}_O}'$ and ${\textbf{V}_S}'$ and maintain the distinction between the untreated video set and support-set. 

In this support-set framework, if the only instruction is that the ${\textbf{V}_O}'$ and $\textbf{T}_{GT}$ are as similar as possible, the encoder will infinitely narrow the distance between these two features, leading to distortions in the $\textbf{V}_{S}$ construction process. Intuitively, we want ${\textbf{V}_O}'$ and $\textbf{T}_{GT}$ to be similar enough in the shared semantic subspace to allow the support-set branch to reflect similar semantic meanings while distinctive in numerical terms to provide extra directions to the model's learning. So a contrastive loss is applied in this part to restrain such behavior. Let $D=1-s({\textbf{v}_O}',{\textbf{v}_S}')$ be the cosine distance between ${\textbf{v}_O}'$ and ${\textbf{v}_S}'$. The loss function can be defined as:
\begin{equation}
\begin{split}
{l_{intra}} &= (1 - Y){D^2} + Y{\{ \max (0,m - D)\} ^2},\\
D &= 1 - s({\textbf{v}_O}^\prime ,{\textbf{v}_S}^\prime ),
\end{split}
\end{equation}
where $m$ denotes the margin for $l_{intra}$, which is also set to 0.2, and the control signal is represented by $Y$. In our settings, we prefer to set the control signal $Y$ close to 1. Our support-set comes from a weighted summation of the original video features in a minibatch. $\textbf{V}_{S}$ is intrinsically strongly linked to $\textbf{V}_{O}$ and is close in the shared semantic subspace. Rather than focusing on the distance between the positive samples and pulling them even closer, we should focus on the negative samples and make them more distinguishable to ensure that the support-set can optimize the encoder parameters truly and effectively. 

\subsection{Decoder}

Our decoder follows the hierarchical structure \cite{zhang2020object} with two LSTMs, including an Attention LSTM and a Language LSTM, noted as deLSTM. For each timestep, we decode the output of the encoder, calculate the final word probability after softmax, and then generate the corresponding word. Captions obtained from the two branches are denoted as $T_{ori}$ and $T_{sup}$. The formula is simplified as follows:
\begin{equation}
\begin{split}
\textbf{T}_{ori}&=deLSTM({\textbf{V}_O}^\prime),\\
\textbf{T}_{sup}&=deLSTM({\textbf{V}_S}^\prime).
\end{split}
\end{equation}

Same as other captioning works, we use Cross-Entropy loss to ensure the correct reasoning of the entire model. The loss function of the captioning process can be defined as:
\begin{equation}
\begin{split}
l_{ori\_cap}&=CrossEntropyLoss(\textbf{T}_{ori},\textbf{T}_{GT}),\\
l_{sup\_cap}&=CrossEntropyLoss(\textbf{T}_{sup},\textbf{T}_{GT}).\\
\end{split}
\end{equation}

In summary, the overall loss can be accomplished as follow:
\begin{align}
l_{overall}&=\lambda_{1}l_{inter}+\lambda_{2}l_{intra}+\lambda_{3}l_{sup\_cap}+l_{ori\_cap},
\end{align}
where $\lambda_{1}$, $\lambda_{2}$, $\lambda_{3}$ are three hyperparameters. 

\section{Performance evaluation}

\begin{table*}[t]
\begin{center}
\caption{The comparison results on MSVD and MSR-VTT. IRv2, Iv3, Res, and 3D-R stand for Inception-ResNet-V2, Inception-V3, ResNet, and 3D ResNeXt-101, respectively. Moreover, B@4, M, R, C denote BLEU-4, METEOR, ROUGE-L, and CIDEr. O/G in the table denotes using extra object/grid features. } 
\vspace{0.15cm}
\label{tab:cap}
\resizebox{0.8\linewidth}{!}{
\begin{tabular}{@{}c|c|c|c|cccc|cccc@{}}
\hline\hline
\multicolumn{1}{c|}{\multirow{2}{*}{Method}} & \multicolumn{1}{c|}{\multirow{2}{*}{Conference}} & \multicolumn{1}{c|}{\multirow{2}{*}{Feature}} & \multicolumn{1}{c|}{\multirow{2}{*}{O/G}} & \multicolumn{4}{c|}{MSVD} & \multicolumn{4}{c}{MSR-VTT} \\
\multicolumn{1}{c|}{} & \multicolumn{1}{c|}{}& \multicolumn{1}{c|}{}& \multicolumn{1}{c|}{}& B@4    & M    & R   & C   & B@4    & M    & R    & C   \\
  \hline\hline
  MGSA\cite{chen2019motion} & AAAI2019 & IRv2+C3D&G& 53.4 & 35.0 & - & 86.7 & 42.4 & 27.6 & - & 47.5 \\

  SAAT\cite{zheng2020syntax} & CVPR2020 & IRv2+C3D&O & 46.5 & 33.5 & 69.4 & 81.0 & 40.5 & 28.2 & 60.9 & 49.1 \\
  RMN\cite{tan2020learning} & IJCAI2020 & IRv2+I3D&O& 54.6 & 36.5 & 73.4 & 94.4 & 42.5 & 28.4 & 61.6 & 49.6 \\
  MGRMP\cite{chen2021motion} & ICCV2021 & IRv2+C3D&G& 53.2 & 35.4 & 73.5 & 90.7 & 42.1 & 28.8 & 61.4 & 50.1 \\
  M$^3$\cite{wang2018m3} & CVPR2018 & VGG+C3D &-& 51.8 & 32.5 & - & - & 38.1 & 26.6 & - & -  \\
  GRU-EVE\cite{aafaq2019spatio} & CVPR2019 & IRv2+C3D &-& 47.9 & 35.0 & 71.5 & 78.1 & 38.3 & 28.4 & 60.7 & 48.1 \\

  \hline\hline
  POS-CG\cite{wang2019controllable} & ICCV2019 & IRv2+I3D  &-& 52.5 & 34.1 & 71.3 & 88.7 & 42.0 & \textbf{28.2} & \textbf{61.6} & 48.7 \\
  Ours & - & IRv2+I3D &-& \textbf{55.4} & \textbf{35.3} & \textbf{72.6} & \textbf{92.9} & \textbf{42.2} & 28.1 & 61.3 & \textbf{49.5} \\
  \hline\hline
  MARN\cite{pei2019memory} & CVPR2019 & Res-101+3D-R &-& 48.6 & 35.1 & 71.9 & 92.2 & 40.4 & 28.1 & 60.7 & 47.1 \\
  MDT\cite{zhao2021multi} & NIPS2021 & Res-101+3D-R &-& 49.0 & 35.3 & 72.2 & 92.5 & 40.2 & 28.2 & 61.1 & 47.3 \\
  SGN\cite{ryu2021semantic} & AAAI2021 & Res-101+3D-R &-& 52.8 & 35.5 & \textbf{72.9} & 94.3 & 40.8 & \textbf{28.3} & 60.8 & 49.5 \\
  Ours & -  & Res-101+3D-R &-& \textbf{55.5} & \textbf{35.6} & 72.6 & \textbf{95.2} & \textbf{41.4} & 28.1 & \textbf{61.4} & \textbf{49.7} \\
  \hline\hline
\end{tabular}
}

\end{center}
\end{table*}
\subsection{Datasets and Metrics}

\textbf{MSVD} \cite{chen2011collecting} consists of 1,970 short video clips taken from Youtube. Each video clip comprises about 41 English descriptions. The dataset is divided into three subgroups to follow previous research: 1,200 clips for training, 100 clips for validation, and 670 clips for testing. 

\noindent \textbf{MSR-VTT} \cite{xu2016msr} is a large-scale dataset for open-domain video captioning. It includes 10,000 video clips from 20 categories, each annotated with 20 English sentences. We employ the standard splits, which are 6,513 clips for training, 497 clips for validation, and 2,990 clips for testing.

We use commonly-used automatic assessment metrics to evaluate the quality of the generated captions, i.e., BLEU-4, METEOR, CIDEr, ROUGE-L. The higher the score, the better the caption quality. 

\subsection{Implementation Details}
\textbf{Feature Extraction.} In our experiments, we use InceptionResNetV2 (IRV2) \cite{szegedy2017inception} as 2D CNN and I3D \cite{kay2017kinetics} as 3D CNN to extract appearance features and motion features respectively, and we equally split 26 clips for each video. For better comparison with the latest work, we also use another feature pair: ResNet-101 \cite{he2016deep} for appearance features, and 3D-ResNext-101 \cite{hara2018can} for motion features.

\noindent  \textbf{Training Details.} We apply teacher-enforced learning (TEL) \cite{zhang2020object} to prompt the caption model to learn the ground-truth word with a certain probability at each training step. Additionally, beam search is applied to produce sentences during the inference phase. Our model is optimized by Adam Optimizer, and the initial learning rate is set to 1e-4 to ensure that the triplet loss converges. The vocabulary is constructed by words with at least 2 occurrences. The hidden size of the LSTM is 1024. The training process lasts 20 epochs. During testing, we use beam search with size 5 for MSVD and size 2 for MSR-VTT.

\subsection{Comparison with State-of-the-Art}
We compare our proposed SMRE with the state-of-the-art methods on the MSVD and MSR-VTT datasets. For a fair comparison, we do not compare our model with methods that utilize additional detectors or grid features. Table 1 shows the comparison results. For both datasets, SMRE outperforms most of the state-of-the-art methods with various backbones, especially on the BLEU and CIDEr metrics. POS-CG \cite{wang2019controllable} and SGN \cite{ryu2021semantic} are chosen as the control groups since they are the latest models with the same features as ours. Compared to POS-CG, our model has improved significantly: 5.5\% in BLEU-4 and 4.7\% in CIDEr on MSVD; 0.7\% in BLEU-4 and 1.8\% in CIDEr on MSR-VTT. Our model outperforms SGN by 5.3\% in BLEU-4 and 1.1\% in CIDEr on MSVD; 1.5\% in BLEU-4 and 0.4\% in CIDEr on MSR-VTT.

\subsection{Ablation Analysis}
In this section, we carry out ablative studies to investigate the contribution of each design in our model on the MSVD dataset. IRv2 and I3D are used to conduct the experiments.

\noindent\textbf{Effects of each module.} Table 2  demonstrates the effectiveness of each component in our proposed model. There are five settings: (1) the baseline model with one encoder-decoder path; (2) support-set structure with the sum of two caption losses $l_{sup}=l_{ori\_cap}+l_{sup\_cap}$; (3) support-set structure and $l_{inter}$; (4) support-set structure and $l_{intra}$; (5) complete structure with four losses. 

From Table 2, we can see that every part of the structure is indispensable. In particular, compared with the baseline model, $l_{sup}$ obtain a slight improvement, which indicates that the support-set can obtain additional supplementary information. To further constrain the relative relationship between raw video features and the support-set, we obtain relative gains by adding $l_{inter}$ and $l_{intra}$ into $l_{sup}$ from inter-modality and intra-modality perspectives, respectively. Moreover, when we combine these three parts with the baseline, the overall model performs best, which shows that the support-set should simultaneously balance the correlation among these perspectives. Thus, the above results conclude that collaborating and constraining with three parts can support the encoder-decoder structure and learn better semantic representations.

\begin{table}[h]
\begin{center}
\caption{Ablation experiments with different combinations of modules. } \label{tab:cap}\vspace{0.25cm}
\resizebox{0.75\linewidth}{!}{
\begin{tabular}{c|c c c c}
  \hline\hline
  \multirow{2}{*}{Function}&\multicolumn{4}{c}{MSVD}\\
  
   & B@4 & M & R & C  \\
  \hline\hline
  baseline&52.3 & 34.4 & 71.0 & 88.6 \\
  \hline
  $l_{sup}$ & 51.2 & 34.6&71.4&89.0 \\
  \hline
  $+l_{inter}$ &55.4&35.1&72.5&90.4\\
  \hline
  $+l_{intra}$&53.8&35.3&72.0&88.7 \\
  \hline
  $l_{overall}$&\textbf{55.4} & \textbf{35.3} & \textbf{72.6} & \textbf{92.9} \\
  \hline\hline
\end{tabular}
}\vspace{-0.4cm}

\end{center}
\end{table}


\noindent\textbf{Effect of $Y$ in $l_{intra}$.}
We further explore the effect of the control signal $Y$ on model performance while intra-modality interacting. The results are summarized in Table 3. From the table, we can find that model performance is better when the control signal $Y$ is closer to 1.0. As mentioned earlier, we speculate the reasons as follows: it is known that when $Y$ is small, $l_{intra}$ focuses more on pulling together positive pairs and vice versa. However, the support-set is inherently similar to the original features. Rather than focusing on the already similar parts, it would be more helpful to focus on the distance pushing to make the model work better. Therefore, we set $Y$=1.0 for MSVD and 0.8 for MSR-VTT.

\begin{table}
\centering
\caption{Effect of the hyperparameter $Y$ in contrastive loss on model performance. } \label{tab:cap}\vspace{0.25cm}
\resizebox{0.75\linewidth}{!}{
\begin{tabular}{@{}c|c c c c@{}}
  \hline\hline
  \multirow{2}{*}{Y}&\multicolumn{4}{c}{MSVD}\\
  
   & B@4 & M & R & C \\
  \hline\hline
  Y=0.0&53.24&35.03&72.19&90.56\\
  \hline
  Y=0.2&53.92&34.34&70.25&87.68\\
  \hline
  Y=0.5&52.95&34.35&70.79&89.17\\
  \hline
  Y=0.8&53.71&34.80&71.56&90.86 \\
  \hline
  Y=1.0&\textbf{55.37}&\textbf{35.34}&\textbf{72.60}&\textbf{92.88}\\
  \hline\hline
\end{tabular}
}\vspace{-0.5cm}
\end{table}

\section{Conclusions}

We present a Support-set based Multi-modal Representation Enhancement (SMRE) model for video captioning to build flexible mapping relationships and mine information in a semantic subspace shared between samples. Specifically, we design a Support-set Construction (SC) module and a Semantic Space Transformation (SST) module to capture subtle connections in multi-modal semantic space, thus gaining better semantic representations. Experimental results on MSVD and MSR-VTT demonstrate the superiority of our SMRE.


\bibliographystyle{IEEEbib}

\bibliography{icme2022template}

\end{document}


\sloppy

\def\x{{\mathbf x}}
\def\L{{\cal L}}

\title{Support-set based Multi-modal Representation Enhancement for Video Captioning}
\name{Supplementary Materials}
\address{}
\maketitle

\section{summary}
Video captioning is a challenging task requiring a thorough comprehension of visual scenes. Existing methods follow a typical one-to-one mapping, which concentrates on a limited sample space while ignoring the intrinsic semantic associations between samples, resulting in rigid and uninformative expressions. In this paper, we present a Support-set based Multi-modal Representation Enhancement (SMRE) model for video captioning to build flexible mapping relationships and mine information in a semantic subspace shared between samples. Specifically, we design a Support-set Construction (SC) module and a Semantic Space Transformation (SST) module to capture subtle connections in multi-modal semantic space, thus gaining better semantic representations. Experimental results on MSVD and MSR-VTT demonstrate the superiority of our SMRE.

\section{ETHICS POLICY}

Our SMRE presents a capacity for generating better captions using semantic relational information among samples. Unlike previous work, our work focus on mining meaningful visual representations and capturing subtle connections in multi-modal semantic space. Our model utilizes existing resources and does not impose much additional burden (computational cost or massive parameters). However, our approach is challenging to employ in realistic scenarios or systems, and our model may be vulnerable, which reminds us of the need to be aware of the potential risks and liability in our approach. Overall, our technique offers an exciting concept, and we hope that this study will spur interest in other multi-modal tasks, prompting them to mine internal data associations and obtain better performance.

\section{Future Work}
In the future, we will continue to explore whether the improvement of the support-set for semantic learning capability can be applied to some large-scale scenarios, such as pre-training models. In addition, the Support-set based Multi-modal Representation Enhancement (SMRE) model proposed in this paper can be used in other multi-modal topics to promote the development of multi-modal semantic representations. 

\section{Other Experiments}
We conduct additional experiments about the size in beam search. We tested different beam sizes in the testing phase, and the results are shown in Tables 1 and Table 2. We found that the increase of beam size on the MSVD dataset can improve the model performance to some extent. The model saturates when the beam size reaches 6. Unlike the MSVD dataset, the excessive beam size on the MSR-VTT dataset has a negative impact on the model performance. Therefore, in our experimental settings, we set beam search with size 5 for MSVD and size 2 for MSR-VTT.

\begin{table}[h]
\caption{Model performance with different beam sizes on MSVD. } 
\label{tab:cap}
\begin{center}
\begin{tabular}{c|cccc}
\hline
\multicolumn{1}{c|}{\multirow{2}{*}{Beam Size}} & \multicolumn{4}{c}{MSVD}      \\
\multicolumn{1}{c|}{}                           & B@4   & M     & R     & C     \\ \hline\hline
2                                               & 52.68 & 35.17 & 72.65 & 91.94 \\ \hline
3                                               & 53.46 & 35.19 & 72.39 & 92.01 \\ \hline
4                                               & 54.54 & 35.40 & 72.54 & 92.48 \\ \hline
5                                               & 55.38 & 35.37 & 72.66 & 92.90 \\ \hline
6                                               & 55.86 & 35.47 & 72.74 & 92.59 \\ \hline
\end{tabular}

\end{center}
\end{table}

\begin{table}[ht]
\caption{Model performance with different beam sizes on MSR-VTT. } 
\label{tab:cap}
\begin{center}
\begin{tabular}{c|cccc}
\hline
\multicolumn{1}{c|}{\multirow{2}{*}{Beam Size}} & \multicolumn{4}{c}{MSR-VTT}   \\
\multicolumn{1}{c|}{}                           & B@4   & M     & R     & C     \\ \hline\hline
2                                               & 41.36 & 28.06 & 61.42 & 49.73 \\ \hline
3                                               & 41.17 & 27.69 & 60.84 & 48.89 \\ \hline
5                                               & 41.55 & 27.58 & 60.89 & 48.88 \\ \hline
\end{tabular}

\end{center}
\end{table}